\pdfoutput=1

\documentclass[11pt]{article}

\usepackage[]{ACL2023}

\usepackage{times}
\usepackage{latexsym}
\usepackage{graphicx}

\usepackage[T3,T1]{fontenc}
\usepackage[noenc]{tipa}

\usepackage[utf8]{inputenc}

\usepackage{microtype}

\usepackage{inconsolata}

\newcommand{\ts}{\textschwa{}}
\newcommand{\eng}{\textipa{N}}

\title{\textit{ManWav}: The First Manchu ASR Model}

   \author{Jean Seo, \ Minha Kang, \ Sungjoo Byun,  Sangah Lee  \\
          Seoul National University \\
         \texttt{\{seemdog, alsgk1123, byunsj, sanalee\}@snu.ac.kr}}

\begin{document}
\maketitle

\begin{abstract}

This study addresses the widening gap in Automatic Speech Recognition (ASR) research between high resource and extremely low resource languages, with a particular focus on Manchu, a critically endangered language. Manchu exemplifies the challenges faced by marginalized linguistic communities in accessing state-of-the-art technologies. In a pioneering effort, we introduce the first-ever Manchu ASR model \textit{ManWav}, leveraging Wav2Vec2-XLSR-53. The results of the first Manchu ASR is promising, especially when trained with our augmented data. Wav2Vec2-XLSR-53 fine-tuned with augmented data demonstrates a 0.02 drop in CER and 0.13 drop in WER compared to the same base model fine-tuned with original data.

\end{abstract}

\section{Introduction}

The landscape of Automatic Speech Recognition (ASR) research has centered around high resource languages such as English. This concentrated attention on high resource languages has deepened the divide between research advancements. While research on English ASR encompasses diverse linguistic variations, including accented and noised speech, the same cannot be said for many low resource languages, though a few basic research including \citet{Chukchi} and \citet{lowASR1} exist. Astonishingly, not a single basic ASR model has been developed for Manchu to date, highlighting a critical void in linguistic inclusivity within the realm of ASR technology. 

The development of a Manchu ASR model holds particular importance in the field of linguistics, as there are no more native speakers of Manchu. Consequently, the available data, whether text or audio, for linguistic study is limited and cannot be replenished. Therefore, it is crucial to maximize the utilization of existing data. However, due to the scarcity of individuals capable of transcribing Manchu audio data, unlabeled data remain unused. If transcribed, this data could prove to be invaluable resource for Manchu research and preservation. Even though the performance of the Manchu ASR system may not be perfect, it would be immensely helpful if it could provide draft transcriptions. This would enable researchers to revise and incorporate them into their studies.

This paper sets out to address the significant gap between high and low resource languages by developing the inaugural Manchu ASR model. This endeavor is underscored by the scarcity of linguistic resources, prompting us to collect all existing Manchu audio data from \citet{KimEtAl2008} in one channel. We try to maximize the cross-lingual capabilities of Wav2Vec2-XLSR-53 \citep{conneau2020unsupervised} by fine-tuning the model with Manchu audio data. The performance of the Manchu ASR model is further enhanced through data augmentation.

The contributions of this study are as follows:
\begin{itemize}
    \item Collecting Manchu audio data in an unified format and correcting corresponding transcriptions
    \item Developing the very first Manchu ASR model with augmented data
    
\end{itemize}

\section{Manchu Language}
The Manchu language, a member of the Tungusic linguistic family, has its roots among the Manchu people of Northeast China and boasts a significant historical role as the official language of the Qing dynasty (1644-1912). Presently, the language confronts a dire state of endangerment, officially denoted a dead language with no more native speakers left.

There have been some efforts to employ technological solutions in the preservation and revitalization of Manchu. These endeavors include the Manchu spell checker \citep{speller}, Manchu-Korean machine translation \citep{seo2023mergen}, and Manchu NER/POS tagging models \citep{lee-etal-2024-manner-manpos}. However, due to the paucity of data, the studies above face challenges and no ASR model has been yet developed.

\section{Data}
\subsection{Materials}
This study leverages Colloquial Manchu data provided by \citet{KimEtAl2008}, in which Colloquial Manchu data is gathered as part of ASK REAL project (Altaic Society of Korea, Researches on Endangered Altaic Languagess \citep{ChoiEtAl2012}). This audio data represents the dialect of Sanjiazi village, located in the Youyi Dowoerzu Manzu Ke'er-kezizu township, Fuyu county, Heilongjiang Province.

The recording took place from February 7th to 14th, 2006 in Qiqihar, Heilongjiang Province, with Mr. Meng Xianxiao (73 years old at that moment). Though Chinese being his first language, Mr. Meng Xianxiao sufficiently served as the speaker, acquiring a comprehensive ability of Manchu by the age of 12.

The data we use in this study is the recordings of the basic conversational expressions and the sentences for grammatical analysis. The length of each recording is 32 minutes and 58 minutes, for a total of 90 minutes. Corresponding transcriptions are basically provided by \citet{KimEtAl2008} and went through some revisions by a Manchu researcher from Seoul National University for better precision. 

\subsection{Transcription}

The phoneme transcription system in this study is based on \citet{KimEtAl2008}. While it shares similarities with the International Phonetic Alphabet (IPA), our system incorporates some distinctions. Specifically, /b, d, g/ represent voiceless unaspirated stops, and /p, t, k/ denote voiceless aspirated stops. Notably, Colloquial Manchu lacks voiced stops, making this transcription system more practical than using diacritic /\textsuperscript{h}/ to indicate aspiration. Next, /\v{j}, \v{c}, \v{s}/ denote voiceless palatal sounds. In IPA system, corresponding sound symbols are [\textctj, \c{c}, \textctc]. But /\v{j}/ is not voiced unlike [\textctj], and /\v{c}/ is the aspirated sound, [\v{c}\textsuperscript{h}]. Some examples can be found in Table~\ref{tab:transcription}.


\begin{table}[ht]
\centering
\resizebox{0.85\columnwidth}{!}{%
\begin{tabular}{cc}
\hline
\textbf{Transcription}   & \textbf{IPA}  \\ \hline
mi\textipa{\ng} \ts{}nj\ts{} bitk s\ts{}w\ts{}.              & mi\textipa{\ng} \ts{}ni\ts{} pitk s\ts{}w\ts{}.   \\
\multicolumn{2}{c}{(Translation: My mother is a teacher.)} \\ \hline
do\v{s}\ts{}n \v{j}o.              & to\textipa{\:z}\ts{}n d\textipa{\:z}o.   \\
\multicolumn{2}{c}{(Translation: Come on in.)} \\ \hline
\end{tabular}%
}
\caption{Examples of our transcription, IPA, and corresponding translation.}
\label{tab:transcription}
\end{table}

\subsection{Data Augmentation}
The scarcity of speech datasets from native Manchu speakers presents a significant challenge, necessitating the adoption of various data augmentation methods. Audio data augmentation methods used to simulate different acoustic environments include:

\begin{itemize}
\item \textbf{Additive noise}: Adding background noise to the audio samples.
\item \textbf{Clipping}: Involves cutting short the audio signals.
\item \textbf{Reverberation}: Applying reverberation effects.
\item \textbf{Time dropout}: Randomly removing segments of the audio.
\end{itemize}

By implementing the above techniques through WavAugment\footnote{https://github.com/facebookresearch/WavAugment} provided by \citet{kharitonov2020data}, we expand the dataset by 100\% respectively, to a total of 400\%, significantly enriching the available train data. Notable is the fact that data augmentation is implemented after the separation of train and test data, ensuring more reliable test results by preventing overlap between the train and test sets. The size of data before and after augmentation is described in Table~\ref{tab:data size}.

\label{sec:DA}

\begin{table}[ht]
\centering
\begin{tabular}{cc}
\hline
\textbf{Before Augmentation} & \textbf{Duration} \\ \hline
train                        & 81 min           \\
test                         & 9.5 min           \\ \hline
\textbf{After Augmentation}  & \textbf{Duration} \\ \hline
train                        & 326.5 min      \\
test                         & 9.5 min                 \\
\hline
\end{tabular}
\caption{The duration of audio files(.wav) in minutes before and after augmentation.}
\label{tab:data size}
\end{table}

\section{Experiment}
\subsection{Models}

Wav2Vec2-XLSR-53 \citep{conneau2020unsupervised} is utilized as the base model. Wav2Vec2-XLSR-53 is a multilingual self-supervised learning (SSL) model from Meta AI\footnote{https://ai.meta.com/} pre-trained with 53 languages. A Wav2Vec2-XLSR-53 model is fine-tuned in two different types of data, leading to two separate fine-tuned models: one with original Manchu data, and the other with augmented Manchu data. We name the model trained with augmented data \textit{ManWav}. The fine-tuning process is conducted through HuggingSound \citep{grosman2022huggingsound}.

\subsection{Experimental Setup}  
Our experiments are conducted using an NVIDIA A100 GPU. We fine-tune our models with learning rate 3e-4, batch size 16, and dropout rate of 0.1. We train Wav2Vec2-XLSR-53 with 400\% augmented data for 1 epoch. On the other hand, Wav2Vec2-XLSR-53 with original data is trained for 5 epochs, ensuring identical train data size for fair comparison.

\section{Result and Discussion}
\subsection{Result}
We use Character Error Rate (CER) and Word Error Rate (WER) as evaluation metrics. CER assesses the accuracy of character transcription, while WER measures the correctness of word recognition. Scores closer to 0 represent better performances in both metrics. WER and CER are the most common and essential metrics in gauging the overall performance of ASR systems.

The experimental results prove the significance of data augmentation in fine-tuning the base model. As depicted in Table~\ref{tab:data augmentation}, using augmented data at the training stage clearly improves the performance, specifically dropping CER by 0.02 and WER by 0.13, indicating the effectiveness using augmented data described in Section \ref{sec:DA}.

Moreover, Table \ref{tab:inference} shows the promising capabilities of \textit{ManWav} in the Manchu speech recognition task. The achieved accuracy is particularly noteworthy given the limited availability of Manchu speech data and considering that Wav2Vec2-XLSR-53 is not initially pre-trained on Manchu.

\begin{table}[ht]
\centering
\begin{tabular}{c|cc}
\hline
\textbf{Data Augmentation} & \textbf{CER}  & \textbf{WER}  \\ \hline
\textbf{before}            & 0.13          & 0.44          \\ \hline
\textbf{after}             & \textbf{0.11} & \textbf{0.31} \\ \hline
\end{tabular}
\caption{The performance of Wav2Vec2-XLSR-53 each trained with data before and after augmentation.}
\label{tab:data augmentation}
\end{table}

\begin{table*}[ht]
\centering
\begin{tabular}{c|c}
\hline
\textbf{Model Prediction}                         & \textbf{Actual Transcription}                     \\ \hline
si jawu\v{c}i bi g\ts{}l jaam si jawu\v{c}i bi g\ts{}l jaam & si jawu\v{c}i bi g\ts{}l jaam si jawu\v{c}i bi g\ts{}l jaam\\
t\ts{}l\ts{} \textcolor{red}{am} \textcolor{red}{dulk\ts{}} ani \ts{}mk\ts{} i\v{c}i bo al\ts{}x\ts{} & t\ts{}l\ts{} \textcolor{blue}{am\ts{}} \textcolor{blue}{dul\ts{}k\ts{}} ani \ts{}mk\ts{} i\v{c}i bo al\ts{}x\ts{} \\
bi sajw\ts{} wak\ts{} bi sajw\ts{} wak\ts{} & bi sajw\ts{} wak\ts{} bi sajw\ts{} wak\ts{} \\
bi sisk\ts{} bitk xolal ba d\ts{} jom mutulko & bi sisk\ts{} bitk xolal ba d\ts{} jom mutulko\\
min do bitk xolal ba joxo & min do bitk xolal ba joxo\\
odun gjak \textcolor{red}{\v{s}axulo} odun gjak \textcolor{red}{\v{s}axulo} & odun gjak \textcolor{blue}{\v{s}awulo} odun gjak \textcolor{blue}{\v{s}awulo} \\ \hline
\end{tabular}
\caption{Examples of inference results from \textit{ManWav}. Wrong predictions are marked red and the corresponding answers are marked blue.}
\label{tab:inference}
\end{table*}

\subsection{Linguistic Analysis}

Taking into account the linguistic characteristics of Manchu, we classify the most common errors in \textit{ManWav} into the following four categories: (1) confusion involving /\ts{}/, (2) confusion and nasalizing of nasal sounds in word-final positions, (3) assimilation between stops, and (4) confusion between /w/ and /x/.

First, there are some uncaptured or mismatched /\ts{}/ sounds in the inference results, particularly in word-final or between sonorants (e.g., /l/) and stops. This occurs because /\ts{}/ can be neutralized with other vowels or even deleted, posing challenges in accurate transcription. As shown in table~\ref{tab:inference}, the locative marker \textit{de} and \textit{am\ts{}} `dad' are sometimes captured as \textit{d} and \textit{am}, indicating apocope of /\ts{}/. The loss of /\ts{}/ is also evident in \textit{dulke}, which originally included /\ts{}/ between the sonorant /l/ and the stop /k/.

Moreover, nasal sounds /n/ and /m/ in word-final positions are frequently overlooked during inference. This could be attributed to the nature of nasal sounds, as they tend to be fused with subsequent vowels, resulting in nasalized vowels, or they may be omitted altogether. The word \textit{gunin} `thought' is an instance of this phenomenon. It is often transcribed as \textit{gunim}, where the final /n/ appears as /m/. The occurrence of nasal stops can sometimes be mistaken for the deletion of the nasalized preceding vowel. For example, the /n/ sound in \textit{ilan} `three' typically nasalizes the following vowels and then is deleted. However, our model erroneously retained the nasal sound in the transcription \textit{ilan}, preserving the final /n/.

Third, the inference results contain pairs that have undergone assimilation based on the articulated position. These pairs were not transcribed as assimilated forms, but this kind of assimilation is a highly productive phenomenon in natural languages. For instance, the /mg/ sequence in \textit{damgu} `tobacco' became /\eng{}g/ in our inference results. This is unsurprising since both /\eng{}/ and /g/ are velar whereas /m/ is bilabial.

Lastly, confusion between intervocalic /w/ and /x/ is frequently observed. To be specific, \textit{\v{s}awulo} `cold' is recognized as \textit{\v{s}axulo} in our model. Given that /w/ is the labial approximant and /x/ is the palatal approximant, it can be noted that these two sounds occupy distinct articulatory positions. However, there is no equivalent unvoiced sound for /w/, and discerning the voicing of approximants becomes challenging when they are in intervocalic positions. 

The above four types of mismatch and corresponding examples are elaborated in Table~\ref{tab:category}.

\begin{table}[ht]
\centering
\fontsize{9.9}{13}\selectfont{
\begin{tabular*}{\columnwidth}{cc}
\hline
\textbf{Mismatch Types} & \textbf{Examples} \\
\hline
(1) \ts{} / \_\_\#, R\_\_C & d\ts{} : d, am\ts{} : am, dul\ts{}k\ts{} : dulk\ts{} \\
(2) n, m / \_\_\# & gunin : gunim, ilan : ila \\
(3) assimilation & damgu : da\eng{}gu \\
(4) w : x / V\_\_V & \v{s}axulo : \v{s}awulo \\
\hline
\end{tabular*}}
\caption{Observed mismatch examples from the inference results written in phonological notations. R refers to sonorants, C consonants, and V vowels. \# means boundary of words; \_\_\# means word-final position.}
\label{tab:category}
\end{table}

\section{Related Work}
\subsection{ASR research in low-resource languages}
There exist some endeavors to apply ASR to low-resource languages. For example, \citet{Chukchi} collect a speech dataset in the Chukchi language and train an XLSR model. Similarly, \citet{Tibetan} improve low-resource Tibetan ASR while \citet{Indigenous} introduce a fully functional ASR system tailored for Seneca, an endangered indigenous language of North America. \citet{Punjabi} propose an effective self-training approach capable of generating accurate pseudo-labels for unlabeled low-resource speech, particularly for the Punjabi language. Furthermore, \citet{lowASR1} explore training strategies for efficient data utilization and \citet{dataaug} investigate data augmentation methods to enhance ASR systems for low-resource scenarios. Other efforts for multilingual ASR or adapting to low-resource scenarios include Kaldi-toolkit\footnote{https://kaldi-asr.org/index.html}, IARPA Babel project\footnote{https://www.iarpa.gov/research-programs/babel}. However, as an extremely endangered language, Manchu has been isolated from all these efforts.

\subsection{Wav2Vec 2.0}

The core innovation of Wav2Vec 2.0 \citep{baevski2020wav2vec} lies in its ability to effectively capture the contextual information in speech through its Transformer-based architecture \citep{vaswani2023attention}. Wav2Vec 2.0 leverages self-supervised training, allowing the training of an ASR model with a minimal amount of labeled data, provided there is an ample supply of unlabeled data. Wav2Vec 2.0 is effective not only in capturing diverse dialects but also in accommodating various languages. XLSR \citep{conneau2020unsupervised} is built on Wav2Vec 2.0 and learns cross-lingual speech representations from raw waveform of speech in multiple languages. XLSR-53 is particularly pretrained on 53 languages, and fine-tuned for Connectionist Temporal Classification(CTC) speech recognition. CTC is a technique used in encoder-only transformer models such as Wav2Vec 2.0, HuBERT \citep{hsu2021hubert} and M-CTC-T \citep{lugosch2022pseudolabeling}.

\section{Conclusion and Future Work}
As an extremely low resource language, Manchu has often been overlooked in linguistic technology. In an effort to maximize the utilization of available Manchu data, the development of an ASR system is essential. We introduce \textit{ManWav}, which involves fine-tuning Wav2Vec2-XLSR-53 on augmented Manchu audio data, with the aim of providing a valuable tool for the study and preservation of Manchu. As the addition of a decoder to an ASR model is known to boost the inference performance \citep{Karita_2019, inproceedings}, enhancing the inference quality with the help of a language model should be studied in the future.

\section*{Limitations}

The primary constraint of this research lies in the scarcity of Manchu audio data. As the audio data used in this research consists only of Colloquial Manchu from one speaker, utilizing \textit{ManWav} in other domains would not show optimized performances, given that ASR models are usually heavily domain-dependent.

\section*{Ethics Statement}

The project paves the way for further innovations in the field and emphasizes the importance of inclusivity in technological advancements, ensuring that the benefits of state-of-the-art technologies are accessible to all linguistic groups, regardless of their resource status. To support further ASR studies on endangered languages, we plan to release \textit{ManWav} in public.

\bibliography{anthology,custom}
\bibliographystyle{acl_natbib}

\end{document}